\documentclass[letterpaper,twocolumn,fleqn]{article} 

\usepackage{ist}
\usepackage{xcolor}

\usepackage{times}
\usepackage{epsfig}
\usepackage{graphicx}
\usepackage{amsmath}
\usepackage{amssymb}
\usepackage{textcomp}
\usepackage{multirow}
\usepackage{adjustbox}
\usepackage{indentfirst}
\usepackage{multirow}
\usepackage{colortbl}
\usepackage{hhline}
\usepackage{subcaption}
\usepackage{enumitem}
\usepackage{verbatim}
\usepackage{tabularx}
\usepackage{booktabs}
\usepackage{float}

\usepackage[symbol]{footmisc}

\usepackage{gensymb}
\usepackage{flushend}
\usepackage{xcolor}
\usepackage{xurl}
\usepackage[colorlinks]{hyperref} 
\hypersetup{
    colorlinks=true,
    breaklinks=true, 
    urlcolor= blue,              
    linkcolor= red, 
    bookmarksopen=false, 
    filecolor=black, 
    citecolor=green, 
    linkbordercolor=blue
}
\usepackage{dirtytalk}

\title{Optical Flow for Autonomous Driving: Applications, Challenges and Improvements}

\author{Shihao Shen$^{\ 1}$, Louis Kerofsky$^{\ 2}$ and Senthil Yogamani$^{\ 3}$ \\ 
$^{1}$Carnegie Mellon University, Pittsburgh, Pennsylvania, U.S. \\
$^{2}$Qualcomm Technologies, Inc., San Diego, California, U.S. \\
$^{3}$Automated Driving, QT Technologies Ireland Limited.}

\date{} 

\begin{document} 

\maketitle 

\thispagestyle{empty} 


\begin{abstract}
Optical flow estimation is a well-studied topic for automated driving applications. Many outstanding optical flow estimation methods have been proposed, but they become erroneous when tested in challenging scenarios that are commonly encountered. Despite the increasing use of fisheye cameras for near-field sensing in automated driving, there is very limited literature on optical flow estimation with strong lens distortion. Thus we propose and evaluate training strategies to improve a learning-based optical flow algorithm by leveraging the only existing fisheye dataset with optical flow ground truth. While trained with synthetic data, the model demonstrates strong capabilities to generalize to real world fisheye data. The other challenge neglected by existing state-of-the-art algorithms is low light. We propose a novel, generic semi-supervised framework that significantly boosts performances of existing methods in such conditions. To the best of our knowledge, this is the first approach that explicitly handles optical flow estimation in low light.
\end{abstract}

\section{INTRODUCTION} \label{sec:Intro}

Advancement in the field of computer vision has enabled the rapid development of perception systems for autonomous vehicles (AV) in recent years. Optical flow estimation, known as the study of how to estimate per-pixel 2D motion between two temporally consecutive frames, is one of the fundamental problems in computer vision that are widely used in autonomous driving. Specifically, optical flow estimation helps vehicles perceive the temporal continuity of the surrounding environment and hence it plays significant roles in time-series-based tasks such as object tracking~\cite{kale2015moving, zhou2018deeptam}, visual odometry~\cite{wang2017deepvo}, semantic segmentation~\cite{rashed2019motion}, motion segmentation~\cite{mohamed2021monocular}, and SLAM systems~\cite{teed2021droid}, to point out a few. Horn and Schunck~\cite{horn1981determining} introduce the first method to compute optical flow through energy minimization and many excellent methods obtain better results based on it. However, the optimizing problem of a complex objective is usually computationally expensive in terms of real-time applications such as AV. To achieve faster and more reliable performance, end-to-end neural networks are proposed~\cite{dosovitskiy2015flownet, sun2018pwc, sun2019models, teed2020raft}. These data-driven learning-based methods are more efficient and robust against challenges, such as occlusions, large displacement and motion blur, that break the brightness constancy and small motion assumptions traditional methods are built upon. Nevertheless, there are still a few unique challenges in AV applications that have been neglected by existing state-of-the-art methods. In this paper, we investigate two commonly encountered challenges among them and propose the solutions respectively: lens distortion and low-light scenes. 

\begin{figure}[t!]
\centering
\includegraphics[width=\columnwidth]{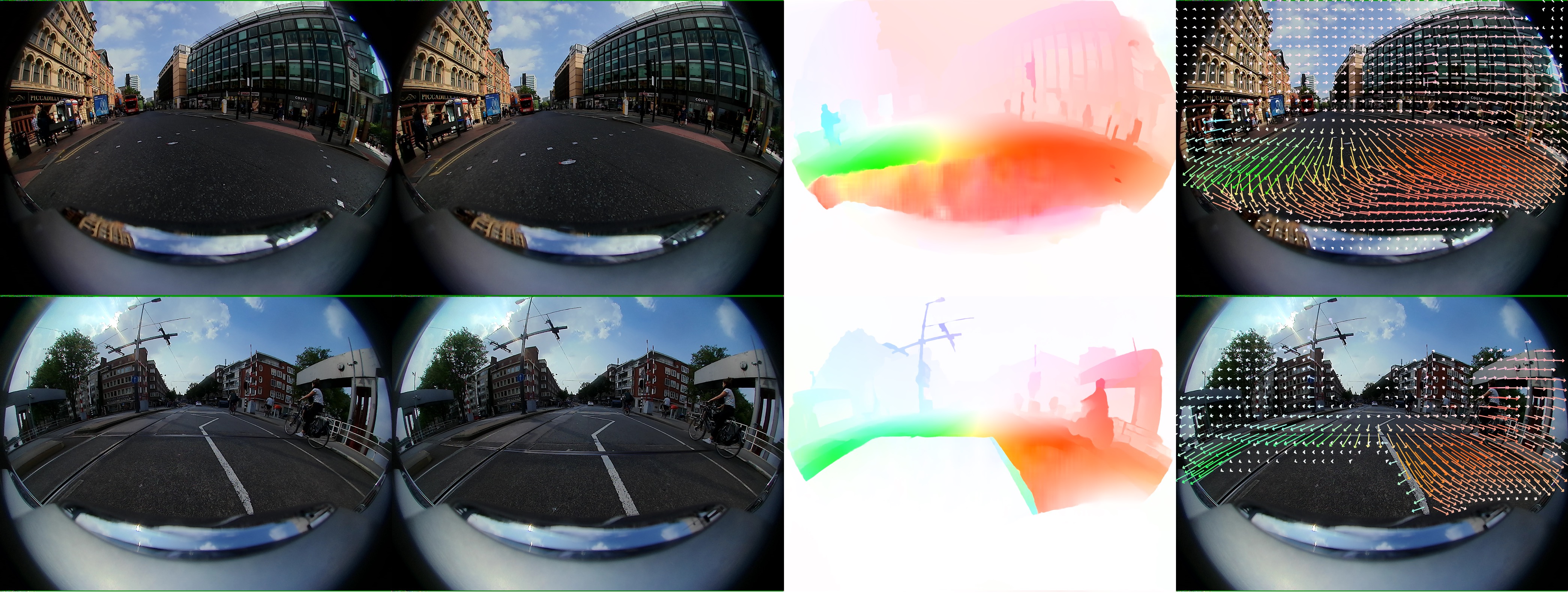}
\vspace*{-2mm}
\caption{Erroneous optical flow estimation by feeding fisheye images into off-the-shelf RAFT~\cite{teed2020raft}. From left to right in each row: current frame, next frame, color coded result, sparse vector overlay plots for better visualization. Note how the estimated flow vectors on the ground are either missing or inconsistent with the vehicle motion.}
\label{fig:fisheye-failure}
\end{figure}

Near-field sensing is a ubiquitous topic for automated driving. Some primary use cases are automated parking systems and traffic jam assistance systems. Near-field sensing is usually achieved by building a surround-view system with a number of wide-angle cameras that come with strong radial distortion. For example, fisheye cameras offer a significantly wider field-of-view (FoV) than standard pinhole cameras, and in practice four fisheye cameras located at the front, rear, and on each wing mirror are sufficient to build a surround-view system for a full-size vehicle~\cite{kumar2022surround}. Although such fisheye systems are widely deployed, to the best of our knowledge, there is no previous work explicitly handling optical flow estimation on images with strong lens distortion, such as fisheye imagery. As shown in Figure~\ref{fig:fisheye-failure}, one of the current state-of-the-art methods~\cite{teed2020raft} shows erroneous results when taking in fisheye images from WoodScape~\cite{yogamani2019woodscape} due to its focus on narrow field-of-view cameras with mild radial distortion only. An intuitive way to solve this is to correct the distortion in the input images as a preprocessing step before passing through the neural network. However, this inevitably leads to reduced field-of-view and resampling distortion artifacts at the periphery~\cite{kumar2021svdistnet}. Without rectification, building an automotive dataset is the major bottleneck in optical flow estimation on fisheye imagery. Very few synthetic datasets provide optical flow ground truth associated with fisheye images~\cite{sekkat2022synwoodscape}, whereas no real-world dataset exists with optical flow ground truth. This is due to the fact that per-pixel motion between every two consecutive frames is extremely difficult to be manually labelled. Simulators~\cite{dosovitskiy2017carla, shah2018airsim} can readily generate background motion but dynamic foreground objects need to be explicitly taken care of. In this paper, we investigate and boost the performance of RAFT on strongly distorted inputs by making use of the only existing dataset with optical flow groundtruth, SynWoodScape~\cite{sekkat2022synwoodscape}.

Most AV applications are expected to operate not only during the day but also at night. Cameras become unreliable and camera-based computations are prone to failure under low-light conditions due to its susceptibility to noise and inconsistent exposure. Alternatively, LiDAR sensors can perform robustly in low-light autonomous driving~\cite{rashed2019fusemodnet} because active sensors that measure the time-of-flight of the emitted lasers are independent of illumination. However, LiDAR is bulky, costly, and requires much more computation as well as memory resources to process the output, which makes it inferior to cameras if the latter can provide equivalently reliable results in low light. Thermal cameras~\cite{dasgupta2022spatio} provide robust low light performance but they are not commonly used in recent automated driving systems. Current optical flow methods show poor capabilities of dealing with low-light data because low light is a complex scenario coming with low signal-to-noise ratio, motion blur and local illumination changes brought by multiple light sources. In addition, current optical flow datasets~\cite{menze2015object, mayer2016large, butler2012naturalistic} are dominated by daytime images. In this paper, we propose a novel, generic architecture that facilitates learning nighttime-robust representations in a semi-supervised manner, without the help of any extra data or sacrificing the daytime performance. To the best of our knowledge, this is the first learning-based method that explicitly handles optical flow estimation in low light. The main contributions of this paper are:
\begin{enumerate}[nosep]
    \item Introduction and investigation of two challenges in optical flow estimation for AV applications: strong lens distortion and low-light scenes.
    \item Implementation and improvement of a baseline optical flow algorithm on fisheye inputs and experimental evaluation.
    \item Implementation of an effective but also generic framework of novel strategies to learn nighttime-robust representations for learning-based optical flow algorithms.
\end{enumerate}

The paper is organized as follows. Section~\ref{sec:related} discusses related work on optical flow estimation in the automotive industry and existing attempts to solve the two aforementioned challenges.
Section~\ref{sec:method} describes the implementation of our proposed flow estimation algorithms for fisheye and low-light inputs respectively, as well as presents the experimental evaluation and results analysis. Finally, Section~\ref{sec:conclu} discusses the remaining challenges for flow estimation in AV applications and concludes the paper.

\section{RELATED WORK} \label{sec:related}


\noindent \textbf{Optical Flow Estimation:} Traditional solutions have been studied and adapted for decades~\cite{horn1981determining, brox2004high}. In order to be robust against more challenging open world problems including lack of features, motions in different scales, and occlusions, recent learning-based methods outperform traditional ones. Dosovitskiy et al.~\cite{dosovitskiy2015flownet} propose FlowNetS and FlowNetC, which is a pioneer work in showing the feasibility of directly estimating optical flow given images. Sun et al.~\cite{sun2018pwc} design PWC-Net, a much more efficient solution based on pyramidal processing, warping and the use of a cost volume. RAFT~\cite{teed2020raft}, proposed by Teed and Deng, demonstrates notable improvement by building multi-scale 4D correlation volumes for all pairs of pixels and iteratively updating flow estimates through refinement module based on gated recurrent units (GRU). All these methods are fully supervised and trained using imagery from a standard pinhole camera. The training data are also collected during the day with sufficient brightness. None of them pays attention to the performance of optical flow in more challenging AV applications such as strong lens distortion and driving at night, which leads to errors and even catastrophic failures. 
\vspace{3pt}

\noindent \textbf{Strong Lens Distortion:} There is very limited work on perception tasks for strongly distorted images such as fisheye images. Popular approaches include rectifying the radial distortion before passing images into any regular perception pipeline. However, this will inevitably bring reduced field-of-view and resampling distortion artifacts especially at the image borders~\cite{kumar2021svdistnet}. Spatially variant distortion that makes closer objects appear larger also poses scaling problems and complexity to geometric perception tasks. Additionally, Rashed et al.~\cite{rashed2021generalized} show that the common use of bounding boxes for object detection no longer fit well for rectangular objects in distorted images. More sophisticated representations for detected objects, such as a curved bounding box exploiting the known radial distortion, are explored in~\cite{rashed2020fisheyeyolo}. Although there is some literature using distorted images without rectification on other perception tasks, such as depth estimation~\cite{kumar2018near, kumar2020fisheyedistancenet}, soiling~\cite{uricar2019desoiling}, visual odometry~\cite{liu2017direct} and multi-task models~\cite{leang2020dynamic, kumar2021omnidet}, there is no previous work estimating optical flow due to the difficulty in labeling ground truth. WoodScape~\cite{yogamani2019woodscape}, KITTI 360~\cite{liao2022kitti} and Oxford RobotCar~\cite{maddern20171} are some well-known autonomous driving datasets containing strongly distorted images such as fisheye images, but none of them has optical flow ground truth. In this paper, we take advantage of the synthetic fisheye dataset published recently, SynWoodScape~\cite{sekkat2022synwoodscape}, which is the first dataset providing optical flow for both foreground and background motions by computing it analytically using other data modalities extracted from the simulator. We train our network using synthetic data from SynWoodScape and evaluate it on real-world fisheye data from WoodScape.
\vspace{3pt}

\begin{figure*}[t]
    \centering
    \includegraphics[width = 2\columnwidth]{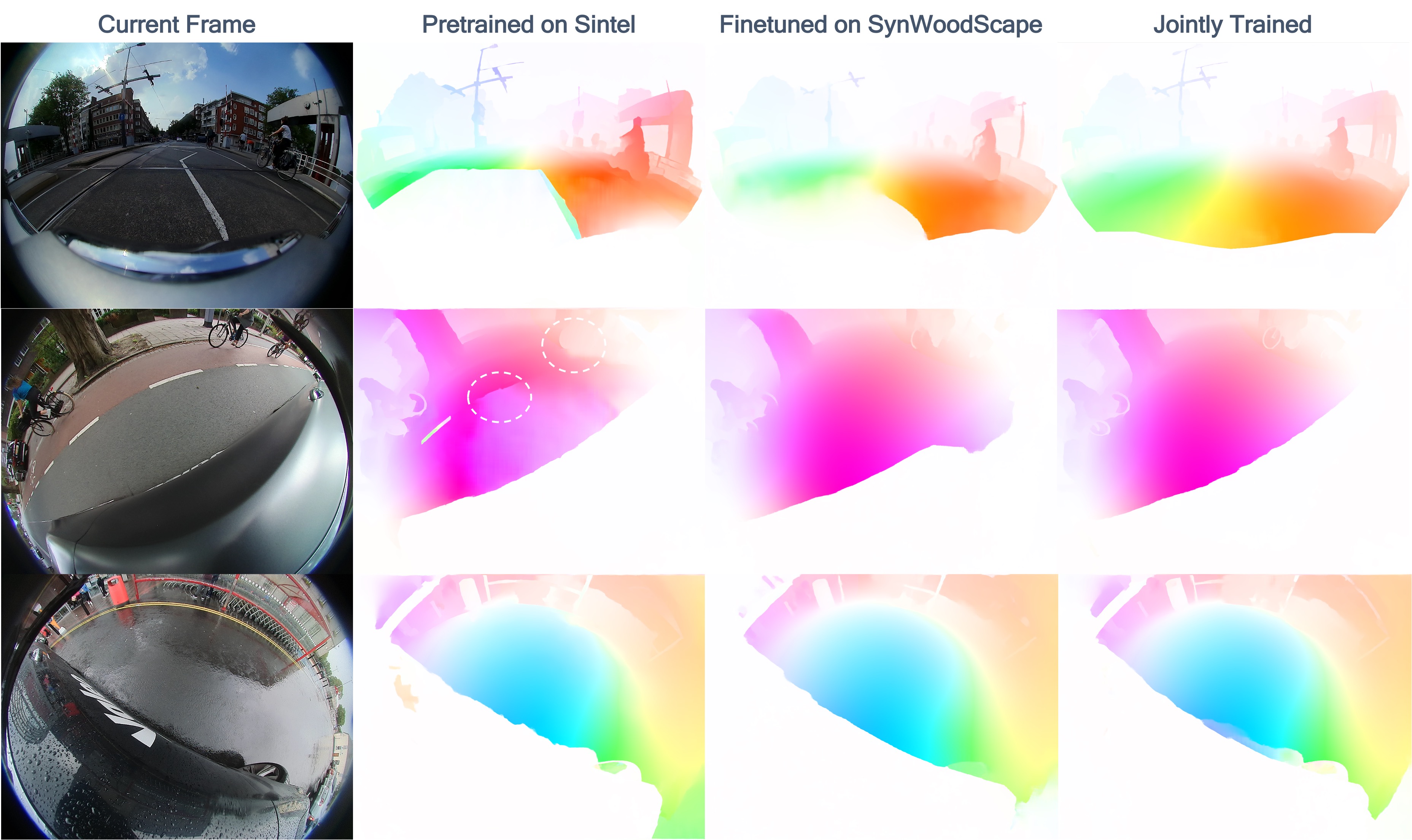}
    \vspace*{1mm}
    \caption{Optical flow estimation (color coded) on real-world automotive data from WoodScape~\cite{yogamani2019woodscape}. Input frames are from the fisheye cameras of front view, right-side view, and left-side view respectively. }
    \label{fig:fisheye}
\end{figure*}

\noindent \textbf{Low-Light Scenes:} Similar to optical flow estimation on strongly distorted images, there is some work handling low light in a few perception tasks~\cite{rashed2019fusemodnet, alismail2016direct, ge2009real} but none of them has proposed an optical flow estimation algorithm that is robust against low-light scenes. Very related to ours, Zheng et al.~\cite{zheng2020optical} propose a method to synthesize low-light optical flow data by simulating the noise model on dark raw images, which is then used to finetune an off-the-shelf network. However, their method is not able to synthesize more realistic characteristics of real-world low-light scenes one would observe in AV applications, such as the motion blur and local illumination changes brought by multiple light sources. Their improvement is also very limited due to the off-the-shelf network is not designed nor trained to learn nighttime-robust representations. In addition, a variety of techniques have been developed for low-light image enhancement~\cite{chen2018learning, lv2018mbllen} and image-to-image translation~\cite{liu2017unsupervised, huang2018multimodal}. The former can preprocess inputs to a flow estimation network during inference by brightening up a given low-light image, whereas the latter can translate a daytime image into its nighttime counterpart so as to complement the lack of optical flow datasets in low light~\cite{rashed2019fusemodnet}. But neither approach facilitates the network training in that the processed data bring in extra complexities such as additional artificial noise, overexposure, or inconsistent image translation across frames. Finally, semi-supervised learning is a common approach to tackling the lack of optical flow data in particular scenarios, where a set of predefined transformations are applied to the original labeled data and the output of the perturbed data are enforced to agree with the outputs of the original data~\cite{laine2016temporal}. For example, Jeong et al.~\cite{jeong2022imposing} use a semi-supervised setup to impose translation and rotation consistency equivariance for optical flow estimation. Yan et al.~\cite{yan2020optical} synthesize foggy images from clean and labelled images in order to avoid flow estimation errors caused in dense foggy scenes. Similar to these semi-supervised methods, we incorporate low-light consistency that facilitates learning explicit nighttime-robust representations without additional labeling.

\section{PROPOSED ALGORITHMS AND RESULTS} \label{sec:method}
In this section, we describe the two proposed optical flow estimation algorithms for strongly distorted inputs and low-light inputs respectively. We also present the corresponding experimental evaluation and results analysis.

\subsection{Strong Lens Distortion}
The limited availability of datasets with strong lens distortion is the bottleneck that prevents recent methods from generalizing to more distorted inputs. With the help of SynWoodScape~\cite{sekkat2022synwoodscape}, the first fisheye dataset providing optical flow ground truth for both foreground and background motions, we are able to train an optical flow model, using RAFT~\cite{teed2020raft} as the backbone, that generalizes well on strongly distorted lenses without sacrificing its original performance on pinhole cameras. 

\begin{figure*}[t!]
    \centering
    \includegraphics[width = \textwidth]{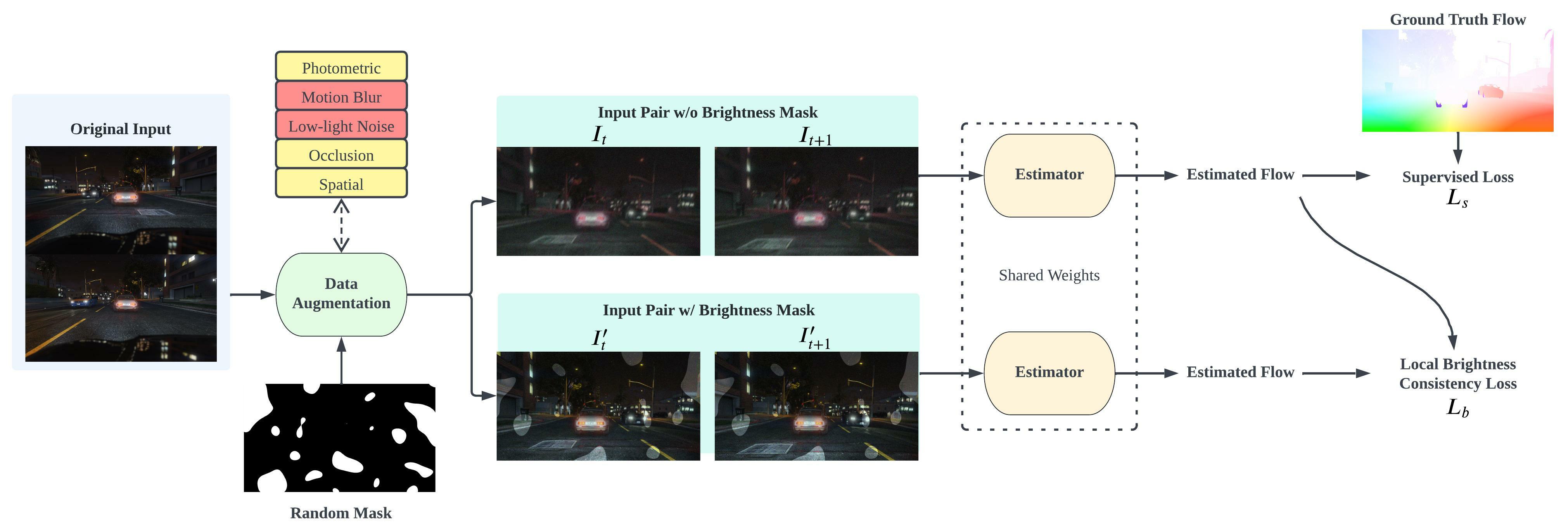}
    \caption{Overview of our proposed framework. During training, the framework takes two consecutive frames as input and passes them through a set of low-light-specific data augmentations as well as applies a random illumination mask. Then the optical flow estimator estimates flow on two pairs of augmented frames in parallel. The network is supervised by two losses: the conventional optical flow loss and the novel brightness consistency loss. During inference, the input frames are directly passed into the estimator which outputs optical flow, as is the standard way in the existing state of the art. }
    \label{fig:architect}
\end{figure*}

We run the off-the-shelf RAFT on real-world fisheye automotive dataset, e.g. WoodScape~\cite{yogamani2019woodscape} and we find sharp and inconsistent optical flow estimation, which is especially illustrated on the ground plane in Figure~\ref{fig:fisheye-failure}. To solve this, we provide two baselines and their qualitative as well as quantitative evaluation. One is to finetune the pretrained RAFT using SynWoodScape, following the training schedule in Table~\ref{subtab:fisheye-ft}. The other is to jointly train RAFT on both SynWoodScape and images from pinhole camera that are regularly used in learning-based optical flow methods~\cite{dosovitskiy2015flownet, menze2015object, mayer2016large, butler2012naturalistic, kondermann2016hci}. The jointly training baseline follows the training schedule in Table~\ref{subtab:fisheye-jt}.

\begin{table}[h]

\vspace*{5mm}
\centering
\caption{Details of the training schedule. Column header abbreviations: LR: learning rate, BS: batch size, WD: weight decay, CS: crop size. Training dataset abbreviations: C: FlyingChairs, W: SynWoodScape, S: Sintel, T: FlyingThings3D, K: KITTI-2015, H: HD1K.}

\vspace*{5mm}

\begin{subtable}{\columnwidth}
\centering
\caption{Finetuning baseline. During the Sintel stage, the dataset distribution is S(.67), T(.12), K(.13), H(.08).}
\label{subtab:fisheye-ft}
\resizebox{\columnwidth}{!}{%
\begin{tabular}{|c|c|c|c|c|c|c|}
\toprule
\textbf{Stage} & \textbf{Weights} & \textbf{Dataset} & \textbf{LR} & \textbf{BS} & \textbf{WD} & \textbf{CS}     \\ 
\midrule
Chairs   & -       & C             & 4e-4          & 6          & 1e-4         & {[}368, 496{]} \\ 
Things   & Chairs  & T             & 1.2e-4        & 3          & 1e-4         & {[}400, 720{]} \\ 
Sintel   & Things  & S+T+K+H       & 1.2e-4        & 3          & 1e-5         & {[}368, 768{]} \\ 
Finetune & Sintel  & W             & 1e-4          & 3          & 1e-5         & {[}600, 800{]} \\ 
\bottomrule
\end{tabular}
}
\end{subtable}

\vspace*{5mm}

\begin{subtable}{\columnwidth}
\centering
\caption{Jointly training baseline. During the Joint stage, the dataset distribution is W(.65), S(.17), T(.13), K(.03), H(.02).}
\vspace*{-2mm}
\label{subtab:fisheye-jt}
\resizebox{\columnwidth}{!}{%
\begin{tabular}{|c|c|c|c|c|c|c|}
\toprule
\textbf{Stage} & \textbf{Weights} & \textbf{Dataset} & \textbf{LR} & \textbf{BS} & \textbf{WD} & \textbf{CS}     \\ 
\midrule
Chairs   & -       & C             & 4e-4          & 6          & 1e-4         & {[}368, 496{]} \\ 
Things   & Chairs  & T             & 1.2e-4        & 3          & 1e-4         & {[}400, 720{]} \\ 
Joint    & Things  & W+S+T+K+H     & 1e-4          & 3          & 1e-5         & {[}368, 768{]} \\ 
\bottomrule
\end{tabular}
}
\end{subtable}

\vspace*{1mm}

\end{table}

We then show the quantitative results in Table~\ref{tab:fisheye_epe}. We use the endpoint error (EPE) as the metric, which is the standard error measure for optical flow estimation. It is the Euclidean distance between the estimated flow vector and the ground truth, averaged over all pixels. We evaluate the two baselines (Stages "Finetune" and "Joint") described above along with the pretrained model (Stage "Sintel") provided by the author on four hold-out test sets from SynWoodScape, Sintel (clean and final passes), and KITTI. SynWoodScape is the only test set of strongly distorted inputs, while the other three assume a pinhole camera model with very little distortion. Although the pretrained model gives outstanding performance on pinhole cameras, its performance significantly drops on fisheye inputs. Our first baseline, the one finetuned on fisheye images, gives the best result on SynWoodScape but has very poor performance on the others. This matches our expectation because both the pretrained and the finetuned models are trained to the best for pinhole camera and fisheye camera respectively, without taking generalization into account. On the other hand, our second baseline, the jointly trained model, keeps the second best while being very close to the best score on all four datasets. Therefore, jointly training provides a straightforward yet strong baseline that generalizes well over lenses with distinct distortions. 

\begin{table}[h]
\centering

\caption{Endpoint-error results on datasets with diverse lens distortion.}
\label{tab:fisheye_epe}

\resizebox{\columnwidth}{!}{%
\begin{tabular}{|c|c|c|c|c|}
\toprule
\textbf{Stage} & \textbf{SynWoodScape} & \textbf{Sintel - Clean} & \textbf{Sintel - Final} & \textbf{KITTI} \\ 
Sintel         & 5.12           & \textbf{1.94}    & \textbf{3.18}    & \textbf{5.10}  \\ 
Finetune       & \textbf{1.40}  & 5.44             & 10.32             & 14.34           \\ 
Joint          & 1.48           & 2.44             & 4.14             & 7.31           \\ 
\bottomrule
\end{tabular}
}

\vspace*{-3mm}

\end{table}

In Figure~\ref{fig:fisheye}, we further show their qualitative results on WoodScape that support the improvements we obtain by jointly training RAFT on a mixture of lens distortions. In the front view case, note how the jointly trained model is able to consistently estimate the flow on the ground as is the major failure of recent methods shown in Figure~\ref{fig:fisheye-failure}. The results on side-view cameras also show the jointly trained model captures finer details than its finetuned counterpart. For example in the right-side view, not only the inconsistency on the ground is solved, but optical flow associated with the bicycle wheel in the upper right corner is also clearly estimated. In the left-side view, the finetuned model misses the flow associated with the vehicle's front wheel, which is captured by the pretrained model, but the jointly trained model "regains" such detailed estimations. In other words, the finetuned model estimates more consistent optical flow, which poses a challenge to the pretrained model due to markedly different projection geometries between fisheye and pinhole cameras, but in return, it loses some details observed by the pretrained model because interesting local features become much less significant given the strong lens distortion. However, the jointly trained model achieves a great trade-off among the previous two: it re-captures the details locally while maintaining good performance globally across different camera views.

\subsection{Low-Light Scenes}
We propose a novel and generic semi-supervised framework that significantly boosts performances of existing state-of-the-art methods in low light conditions. Figure~\ref{fig:architect} shows the architecture of the framework. The benefit of our framework is threefold. First, it is independent from the design of the existing methods, so one can apply it generically to an estimator of his choice (e.g.~\cite{dosovitskiy2015flownet, sun2018pwc, teed2020raft, xu2022gmflow}) and augment its nighttime performance out of the box. Second, semi-supervised learning does not require any extra data as the labeling cost for nighttime optical flow datasets is immense. Lastly, it maintains the estimator's competitive performance on the original daytime data without making any trade-off. 

We first break down the root causes of failures in optical flow estimation under low light and then describe our proposed strategies in the framework that address these root causes accordingly:
\begin{enumerate}[nosep]
    \item the complex noise model of images captured at night,
    \item severe motion blur caused by longer exposure time,
    \item inconsistent local brightness brought by multiple independent light sources in the scene.
\end{enumerate}

Images captured in low light tend to have more complex noises than those captured with sufficient ambient light. Such noises are never synthesized in the data augmentation step by existing methods, which is the first reason why the optical flow estimators fail in low light. Similar to~\cite{zheng2020optical}, we decompose the noise model in low light as an aggregate of the photon shot noise and thermal noise. The former is due to the changing amount of photons hitting the sensor with different exposure levels and pixel locations. The photon shot noise is approximated by a Poisson distribution. Thermal noise refers to the noise in readout circuitry in the sensor and is approximated by a Gaussian distribution. Therefore, we synthesize the low-light noise onto the input frames as one extra data transform in the data augmentation step. Specifically, we sample Poisson and Gaussian parameters, $(a,b)$, from ranges observed in real-world low-light images, formulate it into a single heteroscedastic Gaussian (Equation~\ref{eq:1}), and apply it to an input frame $I$. With probability $0.5$, the low-light noise augmentation is performed on each pair of consecutive frames. 
\begin{equation} \label{eq:1}
\begin{split}
    I(x)=\mathcal{N} \left( \mu=x, \sigma^2=ax+b \right)
\end{split}
\end{equation}

\begin{figure}[t]
    \centering
    \includegraphics[width = \columnwidth]{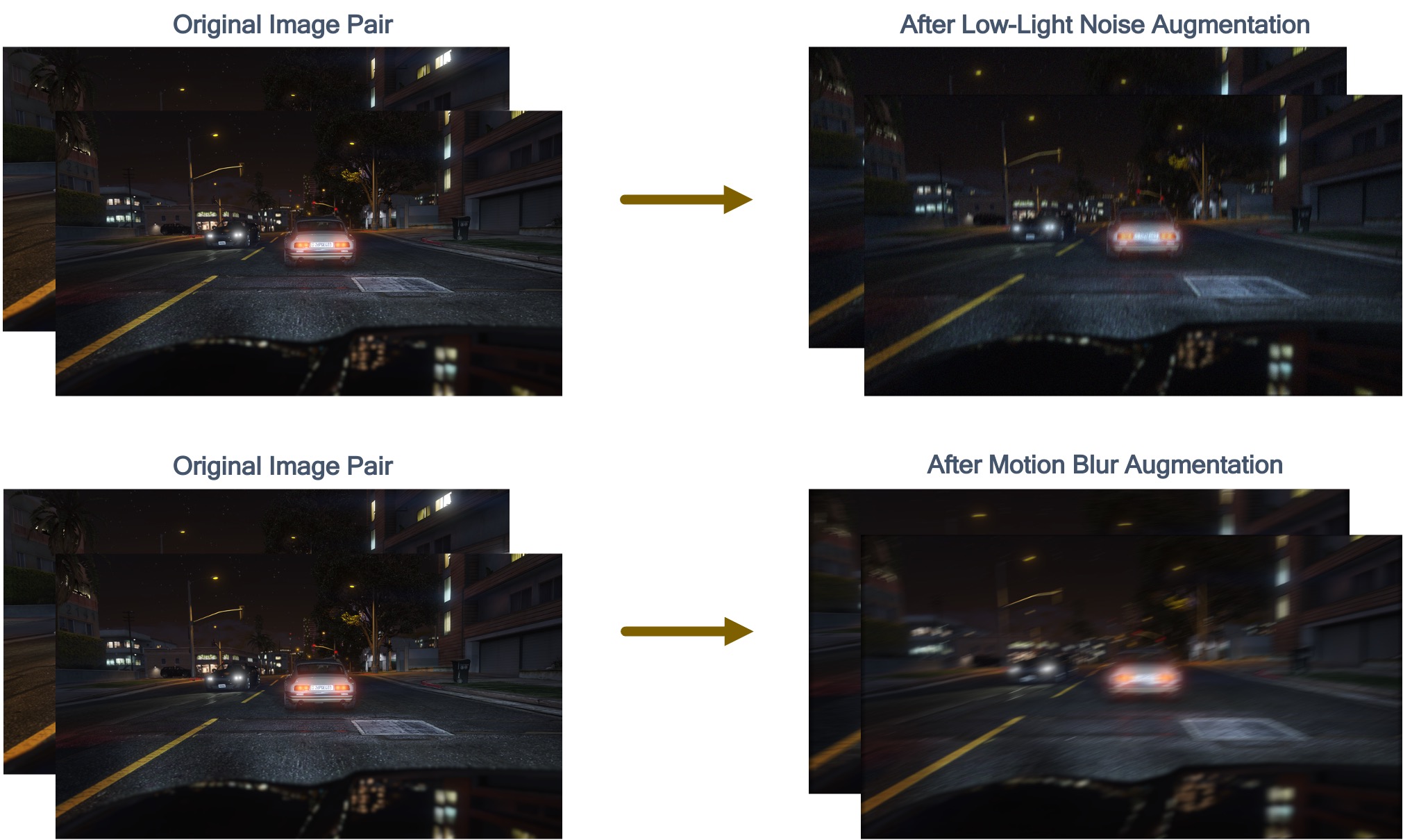}
    \vspace*{-3mm}
    \caption{Effects of low-light noise augmentation and motion blur augmentation.}
    \label{fig:aug}
\end{figure}

Motion blur is another root cause we need to address when estimating optical flow in low light. In order to mimic the blurring effects caused by longer exposure length, we generate authentic motion blur kernels using Point Spread Functions (PSF) at different kernel sizes and intensities. The intensity determines how non-linear and shaken the motion blur looks. Similar to low-light noise, we apply the authentic blurring to a pair of input frames as one extra data augmentation, with probability $0.6$. An illustration of the two introduced data augmentation strategies is shown in Figure~\ref{fig:aug}.

\begin{figure}[t]
    \centering
    \includegraphics[width = \columnwidth]{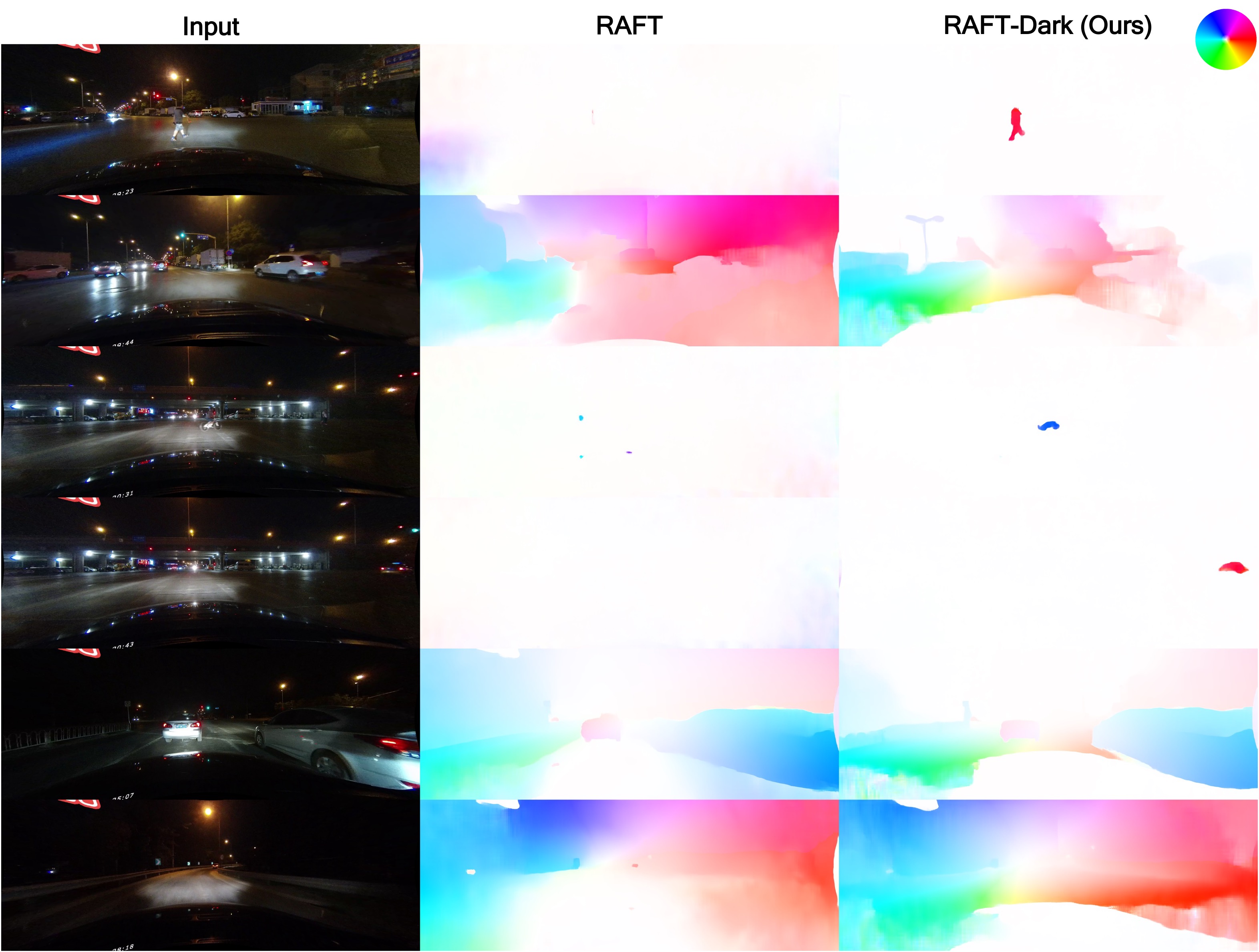}
    \vspace*{-3mm}
    \caption{Optical flow estimation on low-speed sequences from CULane~\cite{pan2018spatial}.}
    \label{fig:raft_dark}
\end{figure}

Inconsistent local brightness is the last but not least root cause. This is due to multiple independent lighting sources existing in a low-light scene (street light, headlight, moonlight, etc.), which leads to uneven bright areas in an image. For example, the ground plane in the original input in Figure~\ref{fig:architect} is illuminated only in front of vehicles' headlights but remains dark elsewhere. Unlike in the daytime where sun is the dominant lighting source, images captured at night have inconsistent local brightness even on the same object. Because optical flow is estimated by matching pixels across two images, such inconsistencies cause existing methods to fail easily. For instance, in Figure~\ref{fig:raft_dark}, the first row shows the catastrophic failure of RAFT when a pedestrian walks from the dark into the vehicle headlight and his illumination changes drastically across frames. In order to resolve this, we resort to semi-supervised learning. Similar to~\cite{jeong2022imposing}, we also adopt the cow-mask~\cite{french2020milking} to create sufficiently random yet locally connected illumination patterns as the inconsistent local brightness occurs in any size, shape and position in images while exhibiting locally explainable structures, depending on the driving environment and the time. We apply the same binary mask to the original pair of input frames and randomly adjust the brightness of pixels according to the mask. The true area of the binary mask is uniformly sampled from 40\% to 70\% of the image. With a probability of $0.5$, we increase the absolute brightness of the true area, whereas in the remaining time we increase the brightness of the false area. Finally, we introduce the local brightness consistency regularization. We use $(I_t', I_{t+1}')$ and $(I_t, I_{t+1})$ to denote the input pair after data augmentation with and without applying a random brightness mask. Both passes in Figure~\ref{fig:architect} are independent except that the spatial transform is shared in order to keep the same cropped areas for consistency loss calculation. The local brightness consistency loss is calculated as follows
\begin{equation}
    L_b = \left \Vert \text{Estimator}\left(I_t, I_{t+1}\right) - \text{Estimator}\left(I_t', I_{t+1}'\right) \right \Vert_2^2 .
\end{equation}
This regularization explicitly constrains the network to output consistent optical flow on $(I_t', I_{t+1}')$ as on $(I_t, I_{t+1})$, which enforces illumination invariance between the estimated optical flow for the original pair the estimated optical flow for the randomly transformed pair. Note how this semi-supervised approach is different from simply adjusting brightness randomly as another data augmentation scheme, which expand training samples without imposition of a sophisticated consistency loss during training. 

We choose RAFT~\cite{teed2020raft} as the estimator and we supervise our network on the aggregated loss $L=L_s+L_b$. $L_s$ is the $l_1$ distance between the predicted flow $\tilde{f}^i(I_t,I_{t+1})$ and ground truth flow $f_{t}$ over all iterations $i$, as in~\cite{teed2020raft}:
\begin{equation}
    L_s = \sum_{i=1}^N \gamma^{N-i} \left\Vert \tilde{f}^i(I_t, I_{t+1}) - f_{t} \right\Vert_1 .
\end{equation}

\begin{figure}[t]
    \centering
    \includegraphics[width = \columnwidth]{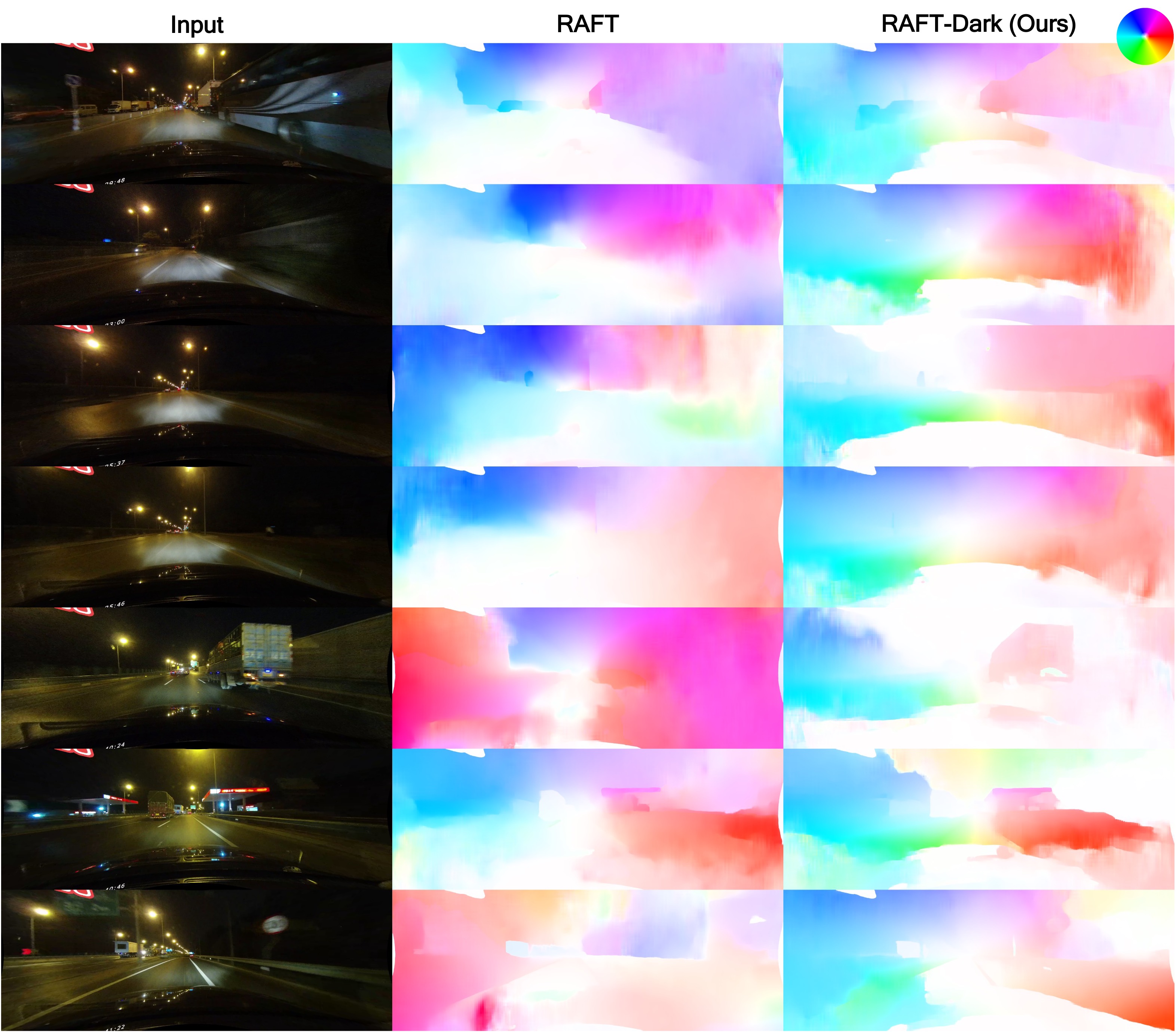}
    \vspace*{-3mm}
    \caption{Optical flow estimation on high-speed sequences from CULane~\cite{pan2018spatial}.}
    \label{fig:raft_dark2}
\end{figure}

Due to the lack of nighttime data with optical flow ground truth, we are restricted to qualitatively evaluating our approach, which we call RAFT-Dark for short. We use CULane~\cite{pan2018spatial}, a large automotive dataset containing a lot of challenging real-world low light sequences. In Figure~\ref{fig:raft_dark}, we show the comparison between vanilla RAFT and RAFT-Dark on some low-speed sequences. RAFT-Dark demonstrates superior performance to RAFT. In the first, third and fourth rows, RAFT-Dark is able to detect motions associated with pedestrians and vehicles that either experience some drastic illumination change or appear to be too dark and noisy. In the other cases, note how RAFT-Dark gives a significantly better estimation on the ground plane as well as the directions and magnitudes that are consistent with the ego vehicle's motion. For convenience, a color coding wheel to visualize per-pixel optical flow vectors is attached to the top right corner: color denotes direction of the flow vector while intensity denotes length of the displacement. Since the ego vehicle always heads forward, the ground truth optical flow vectors in the front camera's image should intuitively point to the image boundaries and away from the image center. And due to the motion parallax, one should expect larger magnitudes of flow vectors toward the image boundaries and small magnitudes around the image center. In other words, although we have no access to numerical ground truth flow, we know the color coded ground truth should exhibit the same pattern as the color wheel: bluish or greenish on the left side while reddish or yellowish on the right side of the image. With this in mind, RAFT fails to estimate correct optical flow consistent with the vehicle's motion, especially in background areas such as the ground plane. On the other hand, RAFT-Dark not only performs well on these areas but also learns to separate the dark sky in some cases and to capture details such as the street light in the second row. Such improvements are further illustrated in high-speed sequences from CULane in Figure~\ref{fig:raft_dark2}. 

Our framework of learning strategies enables RAFT to improve estimation accuracy by more than 50\% on average (based on visual observations), and even solves some catastrophic failures. Although we show our results based on RAFT as the estimator, our framework is generic and one can replace RAFT with any existing state-of-the-art method of one's choice.

\subsection{Discussion}
The goal of this work is to emphasize the importance of addressing optical flow challenges which are not well explored in automated driving. We investigate two of them and propose our solutions accordingly, but the others require further research. Lack of data tends to be the major bottleneck for most data-driven optical flow algorithms. We are able to leverage synthetic data to improve existing methods' adaptation of various lens distortions but the sim-to-real gap still exists when these methods are evaluated on real world fisheye data. Optical flow in low light cannot be addressed in the same way without any synthetic data available. We experiment image enhancement prior to the network inference, but it leads to even worse results because enhancement happens per frame rather than per pair of frames and temporal consistency is easily broken. Without any extra data, our approach takes full advantage of publicly available data and simulates three root causes through novel data augmentation schemes and semi-supervised learning. However, low light is merely one of many scenarios that make optical flow estimation harder. Others include foggy, rainy or snowy weather \cite{dhananjaya2021weather}. A unified and robust approach aiming for all these cases is encouraged and we see it also as an opportunity for further investigation by the community.  

\section{Conclusion} \label{sec:conclu}
Both lens distortion and low light are important problems for higher levels of automated driving, but they are not explored in detail in the optical flow community as there is no public dataset available. Thus we propose our approaches to these two respectively. We implement and improve a state-of-the-art optical flow algorithm by training it on synthetic fisheye data and demonstrating its adaptation to real-world distorted images as well as generalizability over various lens distortions. We implement a novel, generic framework that facilitates learning nighttime-robust representations in a semi-supervised manner, which shows superior performance to the existing state of the art. In future work, we plan to integrate our current solutions into higher-level pipelines as well as explore other unique challenges of optical flow estimation in the context of automated driving.

{\small
\bibliographystyle{ieeetr}
\bibliography{references}
}


 \begin{biography}

\noindent \textbf{Shihao Shen} is a second-year graduate student in the Robotics Institute at Carnegie Mellon University and expects to receive his M.Sc. in Robotic Systems Development in 2023. He worked as an Interim Engineering Intern in the Multimedia Research and Development department at Qualcomm in summer 2022 and this is his work done during his internship. His main research focus is machine learning with applications in computer vision as well as simultaneous localization and mapping (SLAM). \\

\noindent \textbf{Louis Kerofsky} is researcher in video compression, video processing and display.  He received M.S. and Ph.D. degrees in Mathematics from the University of Illinois, Urbana-Champaign (UIUC).  He has over 20 years of experience in research and algorithm development and standardization of video compression.  He has served as an expert in the ITU and ISO video compression standards committees. He is an author of over 40 publications which have over 5000 citations.  He is an inventor on over 130 issued US patents.  He is a senior member of IEEE, member of Society for Information Display.  \\

\noindent \textbf{Senthil Yogamani} is an artificial intelligence architect for autonomous driving and holds a principal engineer position at Qualcomm. He leads the research and design of AI algorithms for various modules of autonomous driving systems. He has over 17 years of experience in computer vision and machine learning including 14 years of experience in industrial automotive systems. He is an author of 110+ publications which have 4000+ citations and 150+ inventions with 85 filed patent families. He serves on the editorial board of various leading IEEE automotive conferences including ITSC and IV and advisory board of various industry consortia including Khronos, Cognitive Vehicles and IS Auto. He is a recipient of the best associate editor award at ITSC 2015 and best paper award at ITST 2012.



 \end{biography}

\end{document}